\documentclass{esannV2}
\usepackage[utf8,latin1]{inputenc}
\usepackage{amssymb,amsmath,array}
\usepackage{graphicx}
\usepackage{hyperref}
\usepackage{subcaption}
\usepackage{wrapfig}
\usepackage{caption}
\begin{document}
\title{Dimensionality of data sets in object detection networks }

\author{Ajay Chawda, Axel Vierling and Karsten Berns
%
%
\vspace{.3cm}\\
%
 TU Kaiserslautern - Informatik \\
Kaiserslautern - Germany
%

}

\maketitle

\begin{abstract}
In recent years, convolutional neural networks (CNNs) are used in a large number of tasks in computer vision. One of them is object detection for autonomous driving. Although CNNs are used widely in many areas, what happens inside the network is still unexplained on many levels. Our goal is to determine the effect of Intrinsic dimension (i.e. minimum number of parameters required to represent data) in different layers on the accuracy of object detection network for augmented data sets. Our investigation determines that there is difference between the representation of normal and augmented data during feature extraction.
\end{abstract}

\section{Introduction and Related work}

\begin{wrapfigure}{r}{0.45\textwidth}
\includegraphics[width=1\linewidth]{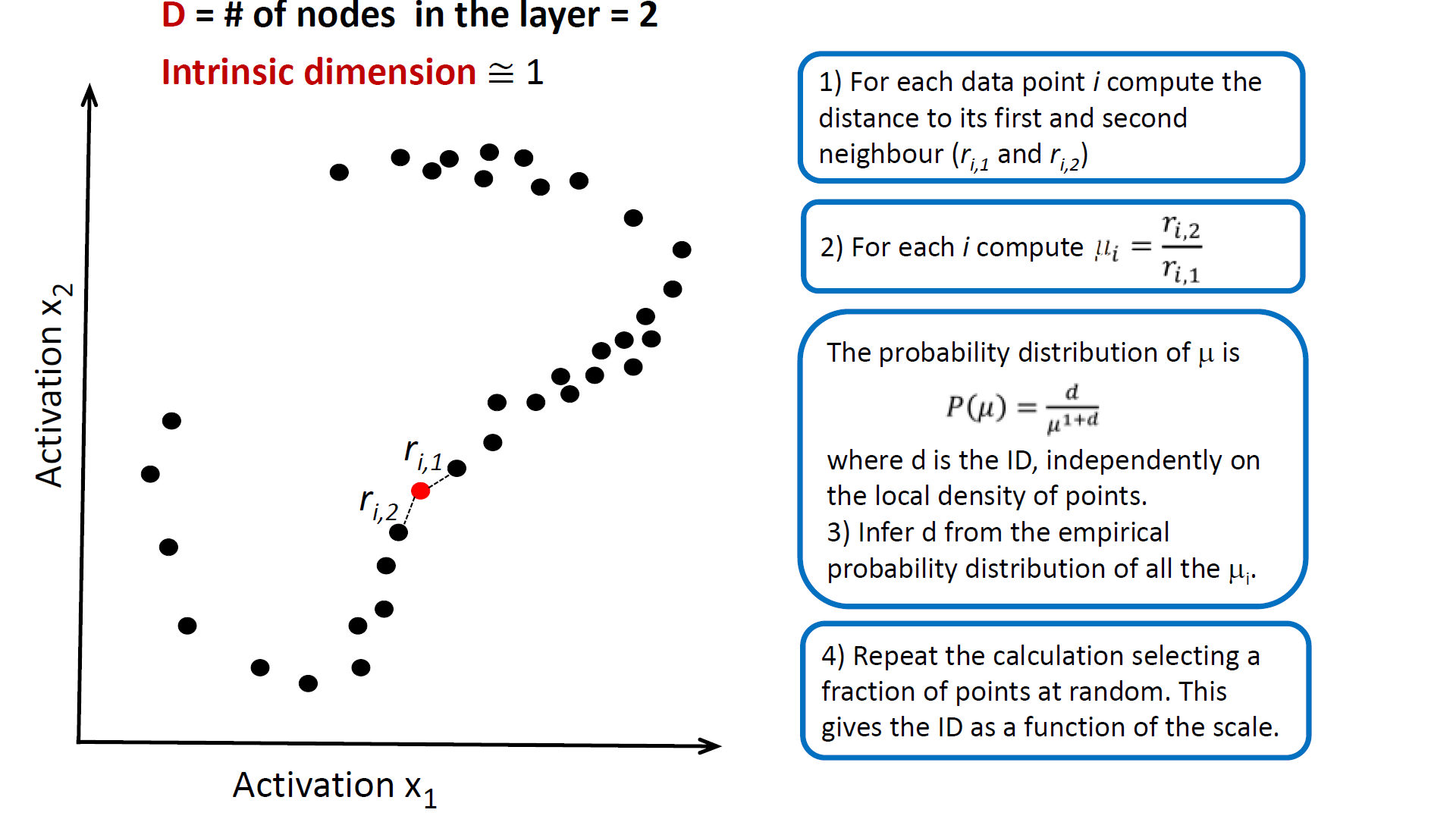} 
\caption{ID is estimated from two nearest neighbour distances. Figure is taken from \cite{ansuini2019intrinsic}.}
\label{fig:1}
\end{wrapfigure}
Autonomous driving is a trending area of research in computer vision. Neural networks are an integral part of the autonomous driving pipeline through which images and lidar points are processed to predict objects. Events have been witnessed where weather changes had led to disastrous consequences of self-driving cars for eg. in 2016 Tesla's self-driving car failed to discriminate between white tractor and bright sky \cite{Tesla2016}. Our objective is to estimate Intrinsic dimension (ID) of augmented data sets in object detection network trained over normal data to observe change in data representation due to noise or affine transformation. Bac et al. \cite{bac2020} states that estimation of ID is important in choosing machine learning methods and its applications including validation, deployment, and explainability. Recognition of labels in intrinsic space is efficient in terms of memory requirements and computation time \cite{gong19}. It is found in \cite{amsaleg2017} that addition of noise to input increases ID. In our study, TwoNN \cite{facco2017} algorithm (Fig \ref{fig:1}) is used to estimate ID. It is based on the ratio of distances between two nearest neighbours which makes it computationally efficient, and also overcomes the issue of data lying on a curved manifold. It is numerically consistent and reliable estimator even in presence of low number of points. From available local and global ID estimators, TwoNN algorithm is used for ID estimation because of interesting results in \cite{ansuini2019intrinsic}. The aim of this paper is to verify \textit{first}, if similar characteristic shape is evident in the case of augmented data sets, \textit{second}, the classification layer ID providing an idea about network performance, \textit{third}, if there is a increase in ID due to irrelevant features and \textit{fourth}, do the augmented data representation behave like an untrained network? The use of three data sets is to study the effect of different data on augmentations. ID is analyzed in Faster R-CNN \cite{ren15fasterrcnn} with VGG-16, VGG-19 \cite{Simonyan15} backbones for KITTI \cite{Geiger2012CVPR}, MS COCO \cite{lin2014microsoft} and VOC \cite{pascal-voc-2007} data sets. Increase in ID due to vertical shift augmentation for KITTI data is observed. Behaviour of rotated images resemble the representation of data in untrained networks for all data sets and dimensional behaviour of COCO data is opposite to KITTI and VOC at classification layer.

\section{Intrinsic Dimension}
One of the geometric properties of representing data in neural network is \textit{Intrinsic dimension} i.e. minimum number of co-ordinates required to represent data without information loss. Local ID estimators \cite{amsaleg:hal-02125331} \cite{houle} compute in local subspaces of data representation. Global ID estimators \cite{facco2017}  compute over whole data point representation. Both global and local ID estimators can be used for estimation in alternate data neighbourhood. Our aim is to estimate ID at different layers for object detection networks and determine the relationship between average precision of augmented data and estimated ID \cite{amsaleg2017}. In \cite{xingjun2018characterizing}, ID characteristics are distinguishable for normal and adversarial generated samples in local space. This motivates us to experiment with ID estimation in global space. TwoNN algorithm is implemented in our paper to estimate ID.
\begin{itemize}
    \item Compute pairwise distances for each point in the data set.
    \item For each point i find two shortest distances $r_1$ and $r_2$ and compute $\mu_i = (r_1 / r_2)$.
    \item Sort the values of $\mu$ in ascending order through a permutation $\sigma$, then define the empirical cumulate $F^{emp}(\mu_{\sigma(i)})$ $\doteq$ $i/N$ .
    \item Fit the points of the plane given by coordinates $\{log(\mu_i), -log(1-F^{emp}(\mu_i)) \}$ with a straight line passing through the origin.
\end{itemize}

The slope of the line give us an estimate of ID. With this approach, the estimated ID is asymptotically correct even for data sampled from non-uniform probability distributions. TwoNN algorithm is referenced from \cite{facco2017}.

\section{Experiments}
 
\begin{table}[htbp]
\centering
  \begin{tabular}{|l|l|c|c|c|}
    \hline
    Dataset & Image Size & Train set & Val set & Network \\
    \hline
    KITTI & 1200 x 1200, & 5500 & 1981 & Faster R-CNN, \\
            & 300 x 300 & & &RetinaNet \cite{Lin2017FocalLF}\\
    \hline
    COCO & 300 x 300 & 80000 & 38000 & Faster R-CNN\\
    \hline
    channel shift& 300 x 300  & 13500 & 3625 & Faster R-CNN \\
    \hline
  \end{tabular}
  \caption{Training and evaluation setup is shown in the above table. During evaluation, complete val set is evaluated and estimating ID with 2000 images for COCO and VOC and 400 images for KITTI from separate test set.}
  \label{table1}
\end{table}

\begin{figure}[h!]
\begin{subfigure}{0.32\textwidth}
\includegraphics[width=4cm, height=3cm]{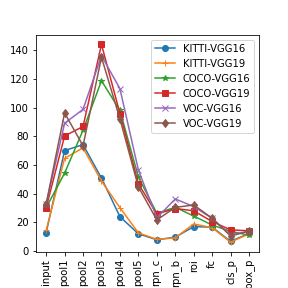} 
\caption{Faster RCNN }
\label{fig:1a}
\end{subfigure}
\begin{subfigure}{0.32\textwidth}
\includegraphics[width=4cm, height=3cm]{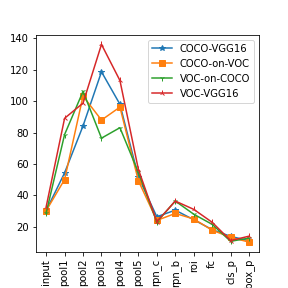} 
\caption{Alternate models}
\label{fig:1b}
\end{subfigure}
\begin{subfigure}{0.32\textwidth}
\includegraphics[width=4cm, height=3cm]{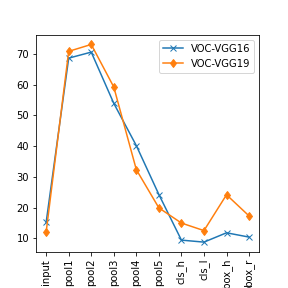}
\caption{RetinaNet}
\label{fig:1c}
\end{subfigure}
\caption{Median intrinsic dimension estimated on KITTI, COCO and VOC data in \ref{fig:1a}, COCO data on VOC trained model and vice-versa in \ref{fig:1b}. Fig \ref{fig:1c} compares RetinaNet on KITTI data for both backbone networks which perform similarly, with  slightly increased ID at bounding box head indicating that using VGG-19 the network generalizes \cite{zhang2016understanding} worse in comparison to VGG-16 backbone at bounding box head layer. }
\end{figure}

In this paper, ID is computed at each pooling layer in VGG backbone network (labeled as pool1, ..., pool5). After feature extraction layers, in Faster R-CNN, ID is computed at classification layer(rpn\_c) and bounding box layer(rpn\_b) in region proposal network. Then, ROI pooling layer(roi), second FC layer(fc) and again for classification(cls\_p) and bounding box layer(box\_p) at the end. In RetinaNet architecture, ID is computed at each pooling layer in VGG backbone. Next layers to compute ID that follow are classification head convolution(cls\_h) block, classification(cls\_l) layer, regression head convolution(box\_h) block and bounding box(box\_r) layer. The reason to compute ID after a block of layers instead of every single layer is due to computational requirements \cite{ansuini2019intrinsic}. ID is also estimated for MS COCO data set on Faster R-CNN model trained on VOC data and alternatively for VOC data set on model trained on COCO data (Fig \ref{fig:1b}). Other implemented augmentations from Table \ref{table2} are \textit{Horizontal Shift}, the image is shifted left or right depending upon the random number generated between -0.7 and 0.7 and \textit{Vertical Shift}, similar to horizontal shift except translation occurs by shifting image top or bottom.
\begin{table}[h!]
\centering
  \begin{tabular}{|l|c|c|c|}
    \hline
    \textbf{Augmentation}  & \textbf{KITTI} & \textbf{VOC} & \textbf{COCO} \\
    \hline
    \textit{normal} - without augmentations & 43.4 & 57.6 & 36.5 \\
    \hline
    \textit{horizontal flip} - flip image on x axis &28.3 & 52.2 & 31.4\\
    \hline
    \textit{channel shift} - shift RGB channels with  &15.1 & 23.6 & 18.3 \\
                random value between -50 and 50 & & &\\
    \hline
    \textit{vertical flip} - flip image on y axis &12.1 & 20.6 & 12.0 \\
    \hline
    \textit{rotation} - image is rotated between 45  &3.6 & 5.2 & 4.3 \\
               and -45 degrees &  &  &\\
    \hline
  \end{tabular}
  \caption{Comparison of average precision(AP) IOU@0.50 over augmented val sets of KITTI, COCO and VOC on Faster R-CNN with VGG-16.}
  \label{table2}
\end{table}

While estimation of ID, bounding box with highest score is used as input to ROI pooling layer from the region proposal network due to constraint of our ID estimation algorithm where each image is represented as a point at layers of our network leads to no change in ID at layers after RPN. What happened while using bounding box with lowest scores? Our results did not have an impact because average precision depends on all objects predicted by the network. Another reason is removal of images from our estimation process if there are no predictions for bounding boxes, because in such scenarios there will be no data points for representation at ROI pooling layer. With 1200 pixels square image the memory requirements during computation of high dimensional tensor (400 x 2304000) is 33.8G. So to reduce computational requirement and save time ID is estimated using 400 images. To check the stability of results, ID is estimated for both small and large sizes, the ID value is higher in the case of a larger image but ID follows similar structure when plotted against the layers used for estimating ID. Plots can be found at our \href{https://github.com/ajaychawda58/ID_CNN}{repository} (https://github.com/ajaychawda58/ID\_CNN).

\section{Results}

\begin{figure}[h!]
\begin{subfigure}{0.32\textwidth}
\includegraphics[width=4cm, height=3cm]{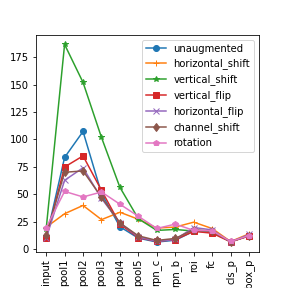} 
\caption{KITTI}
\label{fig:2a}
\end{subfigure}
\begin{subfigure}{0.32\textwidth}
\includegraphics[width=4cm, height=3cm]{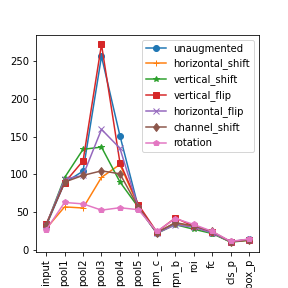} 
\caption{VOC}
\label{fig:2b}
\end{subfigure}
\begin{subfigure}{0.32\textwidth}
\includegraphics[width=4cm, height=3cm]{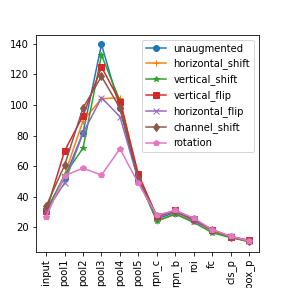}
\caption{MS COCO}
\label{fig:2c}
\end{subfigure}
\caption{Intrinsic dimension estimates over layers of Faster R-CNN on KITTI (\ref{fig:2a}), VOC (\ref{fig:2b}) and COCO (\ref{fig:2c}). Our figures depict variations in estimated ID at backbone layers where features are extracted for RPN and ROI layers. A large shift is evident in \textit{vertical shift} for KITTI data. Rotated images behaviour resembles as represented by an untrained network. }
\label{fig:2}
\end{figure}

As per findings in (See 3.1 in \cite{ansuini2019intrinsic}) for classification tasks  hunchback shape is evident in trained networks whereas in untrained networks the network displays a flat profile. In our experiments, flatter trajectory for rotated images(Fig \ref{fig:2}) is observed, which indicate that rotated images have poor representation in the manifold. It is proven from the evaluation of rotated images where average precision (Table \ref{table1}) is low compared to other augmentations over all data sets. Hunchback profile for other augmented data sets with varying ID at different layers is present in (Fig \ref{fig:2}), hence they are represented better within network in comparison to rotated images.

Vertical shift in KIITI (Fig \ref{fig:2a}) has high ID $\sim$ 187 whereas the normal data has ID $\sim$ 84 at pool1 layer. It may be because of irrelevant features like filling of resized image with interpolation that attribute to increase in ID \cite{ansuini2019intrinsic} and due to original image size of KITTI being around 1200 x 350. When image is shifted vertically and empty pixels are filled by interpolation, the added pixels are irrelevant features to the network. Comparing with COCO and VOC (Fig \ref{fig:2b} \& \ref{fig:2c}) large difference between vertical shift and normal data is absent. Therefore, claim of increased ID can be confirmed due to aspect ratio 3:1 of KITTI images because in case of COCO and VOC the aspect ratio is close to 1:1. If the increase in shift was only due to filling of shifted image, it would be also present in horizontally shifted images. But the absence of increased ID in initial pooling layers for horizontal shift supports our claim.

ID of classification layer does not predict the object detection performance in contradiction to (See 3.2 in  \cite{ansuini2019intrinsic}) that corresponds to relationship between last hidden layer and accuracy of classification. In our case last hidden layers(fc layer) ID also have no relationship with AP (Table \ref{table1}). So, using TwoNN \cite{facco2017} algorithm, dependence of ID with AP over data sets cannot be confirmed but difference in ID is observed at feature extraction level that motivates us to investigate our hypothesis using a different approach later.

While comparing Fig \ref{fig:2a}, Fig \ref{fig:2b} and  Fig \ref{fig:2c} our observations are that ID is lower in classification layer in comparison to bounding box layer for KITTI and VOC data, but for COCO data the phenomenon is reversed with higher ID in classification layer than bounding box layer. One possibility is that the network generalizes poorly at classification layer due to large number of classes(n=91) \cite{ma2018}. Evaluating COCO data on model trained on VOC data and vice-versa, our goal is to investigate how different data sets behave within the network trained on another data set. There is decrease in ID at pool3 layer in both data sets. The reason for decrease can be attributed to change in number of classes in the network affecting the ID at this particular layer because other hyper parameters for the network are same in both data sets. 

\section{Conclusion and Future work}

The presented approach is based on data representation in object detection networks by estimating ID. Results are compared against classification task in \cite{ansuini2019intrinsic} and observed that they are comparable at feature extraction level but not beyond region proposal network. The approach is constrained due to choice of estimator for ID, still interesting behaviour is observed at backbone level, which motivates to continue research with different estimators. Further the research will continue with comparison between current results and model trained on augmentations and network without proposals eg. YOLO\cite{Redmon_2016_CVPR} and eliminate bottleneck due to ID estimation with current approach. Our work starts at a basic level by estimating ID of data sets on Faster R-CNN which indicates the novelty of approach and hope to find more explanations about object detection networks in future.


\begin{footnotesize}


\bibliographystyle{plain}
\bibliography{bibfile}

\begin{thebibliography}{10}

\bibitem{amsaleg2017}
Laurent Amsaleg, James Bailey, Dominique Barbe, Sarah Erfani, Michael~E. Houle,
  Vinh Nguyen, and Milos Radovanovic.
\newblock The vulnerability of learning to adversarial perturbation increases
  with intrinsic dimensionality.
\newblock In {\em 2017 IEEE Workshop on Information Forensics and Security
  (WIFS)}, pages 1--6, 2017.

\bibitem{amsaleg:hal-02125331}
Laurent Amsaleg, Oussama Chelly, Michael~E Houle, Ken-Ichi Kawarabayashi, Milos
  Radovanovic, and Weeris Treeratanajaru.
\newblock {Intrinsic Dimensionality Estimation within Tight Localities}.
\newblock In {\em {Proceedings of the 2019 SIAM International Conference on
  Data Mining}}, pages 181--189. {Society for Industrial and Applied
  Mathematics}, 2019.

\bibitem{ansuini2019intrinsic}
Alessio Ansuini, Alessandro Laio, Jakob~H Macke, and Davide Zoccolan.
\newblock Intrinsic dimension of data representations in deep neural networks.
\newblock {\em arXiv preprint arXiv:1905.12784}, 2019.

\bibitem{bac2020}
Jonathan Bac and Andrei Zinovyev.
\newblock Local intrinsic dimensionality estimators based on concentration of
  measure.
\newblock In {\em 2020 International Joint Conference on Neural Networks
  (IJCNN)}, pages 1--8, 2020.

\bibitem{pascal-voc-2007}
M.~Everingham, L.~Van~Gool, C.~K.~I. Williams, J.~Winn, and A.~Zisserman.
\newblock The {PASCAL} {V}isual {O}bject {C}lasses {C}hallenge 2007 {(VOC2007)}
  {R}esults.

\bibitem{facco2017}
Elena Facco, Maria d`Errico, Alex Rodriguez, and Alessandro Laio.
\newblock Estimating the intrinsic dimension of datasets by a minimal
  neighborhood information.
\newblock {\em Scientific Reports}, 7, 09 2017.

\bibitem{Geiger2012CVPR}
Andreas Geiger, Philip Lenz, and Raquel Urtasun.
\newblock Are we ready for autonomous driving? the kitti vision benchmark
  suite.
\newblock In {\em Conference on Computer Vision and Pattern Recognition
  (CVPR)}, 2012.

\bibitem{gong19}
Sixue Gong, Vishnu~Naresh Boddeti, and Anil~K. Jain.
\newblock On the intrinsic dimensionality of image representations.
\newblock In {\em 2019 IEEE/CVF Conference on Computer Vision and Pattern
  Recognition (CVPR)}, pages 3982--3991, 2019.

\bibitem{houle}
Michael~E Houle.
\newblock Local intrinsic dimensionality i: an extreme-value-theoretic
  foundation for similarity applications.
\newblock In {\em International Conference on Similarity Search and
  Applications}, pages 64--79. Springer, 2017.

\bibitem{Lin2017FocalLF}
Tsung-Yi Lin, Priya Goyal, Ross~B. Girshick, Kaiming He, and Piotr Doll{\'a}r.
\newblock Focal loss for dense object detection.
\newblock {\em 2017 IEEE International Conference on Computer Vision (ICCV)},
  pages 2999--3007, 2017.

\bibitem{lin2014microsoft}
Tsung-Yi Lin, Michael Maire, Serge Belongie, James Hays, Pietro Perona, Deva
  Ramanan, Piotr Dollar, and Larry Zitnick.
\newblock Microsoft coco: Common objects in context.
\newblock In {\em ECCV}, 2014.

\bibitem{ma2018}
Xingjun Ma, Yisen Wang, Michael~E Houle, Shuo Zhou, Sarah Erfani, Shutao Xia,
  Sudanthi Wijewickrema, and James Bailey.
\newblock Dimensionality-driven learning with noisy labels.
\newblock In {\em International Conference on Machine Learning}, pages
  3355--3364. PMLR, 2018.

\bibitem{Redmon_2016_CVPR}
Joseph Redmon, Santosh Divvala, Ross Girshick, and Ali Farhadi.
\newblock You only look once: Unified, real-time object detection.
\newblock In {\em Proceedings of the IEEE Conference on Computer Vision and
  Pattern Recognition (CVPR)}, June 2016.

\bibitem{ren15fasterrcnn}
Ren Shaoqing, He~Kaiming, Girshick Ross, and Sun Jian.
\newblock {Faster R-CNN}: Towards real-time object detection with region
  proposal networks.
\newblock {\em Conference on Computer Vision and Pattern Recognition (CVPR)},
  2015.

\bibitem{Simonyan15}
Karen Simonyan and Andrew Zisserman.
\newblock Very deep convolutional networks for large-scale image recognition.
\newblock In {\em International Conference on Learning Representations}, 2015.

\bibitem{Tesla2016}
The~Tesla Team.
\newblock A tragic loss.
\newblock \url{https://www.tesla.com/blog/tragic-loss}.

\bibitem{xingjun2018characterizing}
Ma~Xingjun, Li~Bo, Wang Yisen, M.~Erfani Sarah, Wijewickrema Sudanthi,
  Schoenebeck Grant, Song Dawn, E.~Houle Michael, and James Bailey.
\newblock Characterizing adversarial subspaces using local intrinsic
  dimensionality.
\newblock In {\em ICLR}, 2018.

\bibitem{zhang2016understanding}
Chiyuan Zhang, Samy Bengio, Moritz Hardt, Benjamin Recht, and Oriol Vinyals.
\newblock Understanding deep learning requires rethinking generalization.
\newblock {\em arXiv preprint arXiv:1611.03530}, 2016.

\end{thebibliography}

\end{footnotesize}


\end{document}